\documentclass{article}

     \PassOptionsToPackage{numbers, compress}{natbib}

     \usepackage[preprint]{neurips_2021}

\usepackage[utf8]{inputenc} 
\usepackage[T1]{fontenc}    
\usepackage{hyperref}       
\usepackage{url}            
\usepackage{booktabs}       
\usepackage{amsfonts}       
\usepackage{nicefrac}       
\usepackage{microtype}      
\usepackage{xcolor}         
\usepackage{microtype}
\usepackage{graphicx}
\usepackage{subfigure}
\usepackage{booktabs} 
\usepackage{bbm}
\usepackage{mathrsfs}
\usepackage{amsmath}
\usepackage{amssymb}
\newtheorem{proposition}{Proposition}

\usepackage{xcolor}
\usepackage{framed}
\usepackage{multirow}
\usepackage{pifont}

\title{A Game-Theoretic Taxonomy of Visual Concepts \\in DNNs}

\author{%
Xu Cheng \\
  Shanghai Jiao Tong University\\
  \texttt{xcheng8@sjtu.edu.cn} \\
   \And
   Chuntung Chu\\
   University of Toronto \\
   \texttt{chuntung.chu@mail.utoronto.ca} \\
   \AND
   Yi Zheng\\
  China University of Mining \& Technology \\
   \texttt{zqt1900202051g@student.cumtb.edu.cn} \\
   \And
   Jie Ren \\
   Shanghai Jiao Tong University \\
   \texttt{ariesrj@sjtu.edu.cn} \\
   \And
   Quanshi Zhang\thanks{Correspondence.} \\
   Shanghai Jiao Tong University \\
   \texttt{zqs1022@sjtu.edu.cn} \\
}

\begin{document}

\maketitle

\begin{abstract}
In this paper, we rethink how a DNN encodes visual concepts of different complexities from a new perspective, \emph{i.e.} the game-theoretic multi-order interactions between pixels in an image.
Beyond the categorical taxonomy of objects and the cognitive taxonomy of textures and shapes, we provide a new taxonomy of visual concepts, which helps us interpret the encoding of shapes and textures, in terms of concept complexities.
In this way, based on multi-order interactions, we find three distinctive signal-processing behaviors of DNNs encoding textures. Besides, we also discover the flexibility for a DNN to encode shapes is lower than the flexibility of encoding textures. Furthermore, we analyze how DNNs encode outlier samples, and explore the impacts of network architectures on interactions. 
Additionally, we clarify the crucial role of the multi-order interactions in real-world applications.
\emph{The code will be released when the paper is accepted.}
\end{abstract}

\section{Introduction}
Although deep neural networks (DNNs) have achieved significant success in various applications, they are still considered as black boxes. Previous post-hoc explanations of visual concepts encoded by DNNs mainly visualize network features \cite{zeiler2014visualizing,yosinski2015understanding,mahendran2015understanding,dosovitskiy2016inverting} or extract correlations between input pixels and network features~\cite{Bau2017cvpr,kim2017interpretability,fong2018net2vec,islam2021shape}. Recently, \citet{li2021shapetexture} proposed a method to measure whether a model mainly encoded textures or shapes for classification.
These efforts focused on the distribution of visual concepts encoded over different filters/layers in an experimental manner, by observing correlations between features and the annotated concepts.
 
Beyond the categorical taxonomy of objects and the cognitive taxonomy of textures and shapes, in this study, we provide a different but more essential perspective to interpret signal-processing behaviors of DNNs encoding visual concepts. 
Specifically, we classify visual concepts into simple (local) concepts, middle-complex concepts, and complex (global) concepts, and further analyze distinctive properties of encoding each type of concepts.
Thus, the new taxonomy of concepts helps us rethink and interpret the encoding of shapes and textures in a more fine-grained manner.
 
\begin{figure}[t]
	\centering
	\includegraphics[width=\linewidth]{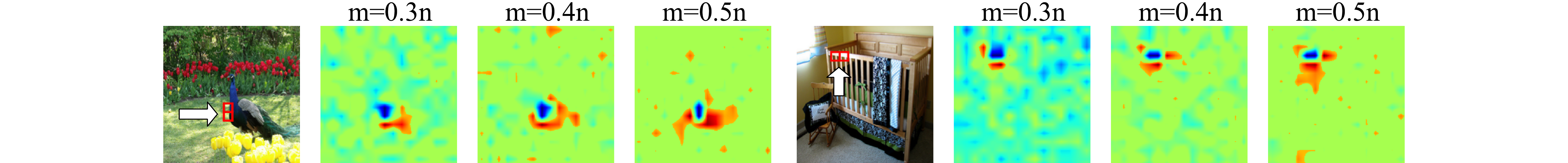}
	\vspace{-15pt}
	\caption{Contexts of the $m$-order interaction {\small$I^{(m)}(i,j)$}. Pixels $(i,j)$ were indicated by white arrows. Pixels which significantly collaborated with $(i,j)$ were regarded as a context, as shown in heatmaps. We simply extended\protect\footnotemark[1] the method in \cite{zhang2020interpreting} to visualize salient concepts of the multi-order interaction.}
	\label{visualization}
	\vspace{-15pt}
\end{figure}

To this end, we use multi-order interactions~\cite{zhang2021interpreting} between input pixels in the game theory to represent concepts of different complexities. 
The interaction between two input pixels measures the marginal attribution 
from the collaboration between these two pixels, in comparison with the case they work individually.
Then, the multi-order interaction represents the additional attribution brought by collaborations among these two pixels and 
other involved contextual pixels, where the order refers to the number of contextual pixels in the collaboration.
As Fig.~\ref{visualization} shows, low-order interactions usually represent relatively simple and local concepts. In comparison, high-order interactions mainly indicate relatively global and complex concepts. 

Based on multi-order interactions, we discover that the DNN usually encodes textures in three different ways, \emph{i.e.} local textures, middle-complex textures, and global textures, respectively.
1) Textures encoded in low-order interactions mainly represent the common distribution of local colors, which can be generalized to different patches of the same texture category. 
2) The DNN may also encode textures using middle-order interactions, and the encoding of such textures is closer to the memory of specific middle-complex instances. 
As Fig.~\ref{texture_class_result} shows, in the fine-grained texture classification, low-order interactions usually encode common, and widely-shared local textures. In comparison, middle-order interactions mainly model the subtle difference between similar textures.
3) Textures encoded in high-order interactions mainly represent global textures, which frequently appear in training samples 
(\emph{e.g.} the sky/ocean/grassland).

Furthermore, the low-order and middle-order interactions reflect the difference in flexibility between encoding textures and encoding shapes. The DNN may flexibly encode a specific texture as either plenty of repetitive local and simple textures corresponding to low-order interactions, or a few middle-complex textures of middle-order interactions. 
For example, the texture of the marble in Fig.~\ref{noise} is made up of repetitive local textures. Hence, this texture can be modeled as either the ensemble of massive local textures or the ensemble of a few middle-complex textures.
In comparison, the encoding of shapes is less flexible. \emph{I.e.} each shape has a relatively robust distribution of interactions through different orders. 
For example, the DNN usually encodes a bicycle as the ensemble of middle-complex shapes such as the frame in Fig.~\ref{noise}, instead of massive local shapes.
Particularly, when the DNN mainly uses high-order interactions to represent shapes, these interactions mainly reflect the memory of specific large-scale shapes or outliers.

Besides, we analyze how the kernel size and the layer width of ResNets influence multi-order interactions. We find that the ResNet with a large kernel size and a wide ResNet are likely to learn a few discriminative interactions towards a specific category. In comparison, the ResNet with a small kernel size and a narrow ResNet tend to learn plenty of widely-shared, generic, but less discriminative interactions.
Furthermore, we clarify the crucial role of multi-order interactions in boosting the classification performance, and explaining and alleviating adversarial effects.

Contributions of this paper are summarized as follows.
1) We provide a new perspective to re-classify visual concepts, which helps us interpret how DNNs encode textures, shapes, and outlier samples in a fine-grained manner.
2) We also analyze the impacts of network architectures on interactions.

\footnotetext[1]{Technical details of visualizing contexts of multi-order interactions were shown in the supplementary material. Please see the supplementary material for more results.}
\section{Related Work}
We limit the discussion within the scope of post-hoc explanations of trained DNNs for visual tasks.

\textbf{Semantic explanations for DNNs.}
 An intuitive way to interpret DNNs is to visualize visual concepts encoded in intermediate layers of DNNs \citep{erhan2009visualizing,hinton2012practical,zeiler2014visualizing,yosinski2015understanding,mahendran2015understanding,dosovitskiy2016inverting,nguyen2016synthesizing,pmlr-v48-oord16,simonyan2017deep,greydanus2018visualizing,islam2021shape}. Moreover, some studies estimated the pixel-wise or regional attribution/importance/saliency of an input image to the network output~\citep{zhou2016learning,ribeiro2016should,selvaraju2017grad,kindermans2017learning,fong2017interpretable,ribeiro2018anchors,chattopadhay2018grad,lundberg2017unified,kindermans2018learning,dhurandhar2018explanations,Kapishnikov_2019_ICCV,hooker2019benchmark,dombrowski2019explanations,schulz2020restricting}.

The dissection of visual concepts is also a classical direction of interpreting DNNs.~\citet{Bau2017cvpr} manually annotated six types of visual concepts (\emph{objects, parts, scenes, textures, materials} and \emph{colors}), and associated units of DNN feature maps with these concepts. Furthermore, the first two concepts could be summarized as shapes, and the last four concepts could be summarized as textures~\cite{Zhang2018cvpr}.~\citet{fong2018net2vec} used a set of filters to represent a single semantic concept, and built up many-to-many projections between concepts and filters. TCAV~\citep{kim2017interpretability} measured the importance of user-defined concepts to classification results.

In summary, previous semantic dissection was mainly conducted in an experimental manner, \emph{i.e.} people associated filters of a DNN with manually annotated concepts. In contrast, we propose to explain the essence of signal processing for different concepts, instead of simply computing the distribution of concepts via experimental observations. 


\section{Analyzing and improving visual concepts of different complexities}
\label{algorithm}
In this section, we use multi-order interactions to explain the essence of signal processing for concepts of different complexities. This explanation helps us rethink how DNNs encode textures and shapes in a fine-grained manner.

\textbf{Preliminary: multi-order interactions.}
Because input pixels of an image do not contribute to the network prediction independently, there often exist interactions between different pixels. Given a trained DNN $v$ and an image $x$ with $n$ pixels, $N=\{1,2,\cdots,n\}$, the Shapley interaction index~\cite{Grabisch1999AnAA} has been proposed as a standard metric for the interaction between pixels.
The Shapley interaction index $I(i,j)$ measures whether the absence/presence of the pixel $i$ can change the attribution of the pixel $j$.
Compared to the Shapley interaction index, in this paper, we introduce multi-order interactions~\cite{zhang2021interpreting}, and illustrate that $I(i,j)$ can be decomposed as the sum of multi-order interactions.
The multi-order interaction $I^{(m)}(i,j)$ measures the additional attribution brought by collaborations among $i$, $j$ and other $m$ contextual pixels, in a more fine-grained manner.
Please see the supplementary material for more details about the meaning and utility of both the Shapley interaction index and the multi-order interaction.
\begin{equation}
\label{eqn:multi-order_interaction}
I(i,j) =\frac{1}{n-1} \sum\nolimits_{m=0}^{n-2} I^{(m)}(i,j), \quad
I^{(m)}(i,j) = \mathbb{E}_{S\subseteq N\backslash \{i,j\},|S|=m}[\Delta v(S,i,j)].
\end{equation}
Here, $\Delta v(S,i,j) \overset{\textrm{def}}{=} v(S \cup \{i,j\})-v(S \cup \{i\}) - v(S\cup \{j\}) + v(S)$, and $v(S)$ represents the network output when we keep pixels in $S$ unchanged and mask pixels in $N\setminus S$ by following~\cite{ancona2019explaining}.
If $I^{(m)}(i,j) > 0$, the presence of $j$ will increase the attribution of $i$. Thus, we consider pixels $i$ and $j$ have positive interactions.
Similarly, If $I^{(m)}(i,j) <0$, $i$ and $j$ conflict with each other, thereby having negative interactions. Besides, if $I^{(m)}(i,j) \approx 0$, then $i$ and $j$ are independent.

Moreover, as Fig.~\ref{visualization} shows, the order $m$ represents contextual complexity of the interaction. For a low order $m$, $I^{(m)}(i,j)$ measures the interaction between $i$ and $j$ \emph{w.r.t.} simple contextual collaborations with a few pixels. Whereas, for a high order $m$, $I^{(m)}(i,j)$ refers to the interaction between $i$ and $j$ \emph{w.r.t.} complex contextual collaborations with massive pixels.

In addition, we have proven that $I^{(m)}(i,j)$ satisfies \emph{linearity, nullity, symmetry, marginal attribution, accumulation,} and \emph{efficiency} properties, which reflect the trustworthiness of the $m$-order interaction\footnote[2]{Please see the supplementary material for more details.}.

\subsection{Concepts of low/middle-order interactions}
 \begin{proposition} 
\label{pro1}
Low-order interactions mainly reflect simple, common, and elementary visual concepts without encoding rich contexts. Whereas, middle-order interactions mainly represent complex textural/shape information.
\end{proposition}
\vspace{-5pt}

In order to verify Proposition~\ref{pro1}, we visualized interaction contexts of different orders.
Fig.~\ref{visualization} shows that low-order interactions mainly reflected local edges, local colors, or local textures. In comparison, middle-order interactions mainly represented middle-scale textures/shapes, which were more complex than low-order interactions.

$\bullet\quad$\textbf{Understanding 1-semantics.} \emph{The modeling of local textures can be roughly divided into two types without clear boundaries. 1) Textural information encoded in low-order interactions mainly represents the simple and common color distribution, which can be generalized to patches of similar textures. 2) In comparison, textural information encoded in middle-order interactions tends to be complex and hard memorized as specific middle-complex instances by DNNs. Moreover, textures are more likely to be encoded as low-order interactions than middle-order interactions, unless low-order interactions are not discriminative enough for inference.}

\begin{figure}[t]
	\centering
	\includegraphics[width=\linewidth]{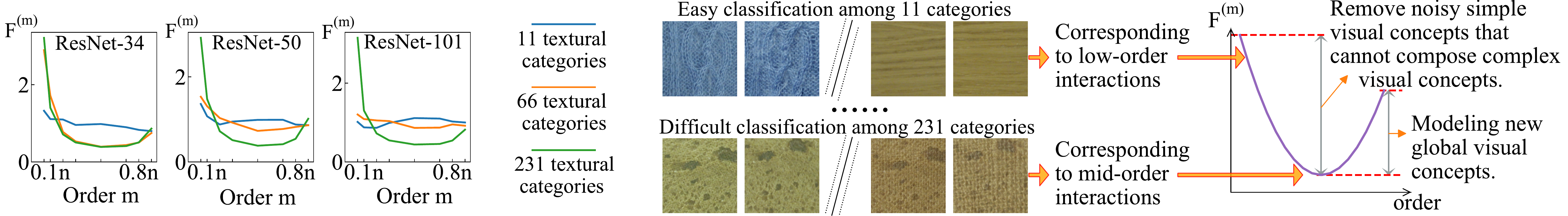}
	\vspace{-18pt}
	\caption{The $m$-order normalized interaction strength {\small${F^{(m)}_{\textrm{strength}}}$} learned for different numbers of texture categories, $0.05n \leq m \leq 0.9n$.}
	\label{texture_class_result}
	\vspace{-6pt}
\end{figure}

\textbf{\emph{Verification}.}
Here, we constructed different datasets to force DNNs to learn textures of different complexities. In this way, we could verify the correlation between texture complexity and the interaction order. As Fig.~\ref{texture_class_result} (right) shows, compared to classifying a few textures, the classification of massive fine-grained textures usually forced a DNN to learn more subtle differences between similar textures, and to remove all ambiguous collaborations. Thus, only a few and less diverse middle-complex collaborations subtly distinguishing fine-grained textures could be encoded. To this end, we used an interaction strength metric to verify whether the fine-grained texture classification made the DNN encode fewer middle-order interactions.

Specifically, we constructed three texture datasets. The first was the original KTH-TIPS-2b dataset \cite{caputo2005class}, which contained $11$ texture categories. To construct the second dataset, we merged images from each pair of texture categories in the first dataset to generate a new texture category, thereby obtaining  {\small$11+\tbinom{11}{2} = 66$} texture categories in total. Similarly, to construct the third dataset, we combined images from three texture categories in the first dataset to generate a new texture category, and achieved overall {\small$11+\tbinom{11}{2}+\tbinom{11}{3} =231$} texture categories. We used these three datasets to train three versions of each ResNet-34/50/101~\cite{he2016deep} model. The $m$-order interaction strength was quantified as
\begin{equation}
\label{eqn:significance}
I^{(m)}_{\textrm{strength}}=\mathbb{E}_{x \in \Omega}\Big[\mathbb{E}_{ i,j}[|I^{(m)}(i,j|x)|]\Big],
\end{equation}
where $\Omega \subseteq \mathbb{R}^{n}$ represented the set of training samples. To reduce the computational cost, $I^{(m)}(i,j|x)$ was calculated at the grid level by dividing the image into $32 \times 32$ grids, thereby $n = 1024$.

Fig.~\ref{texture_class_result} shows the $m$-order normalized interaction strength {\small$F^{(m)} = I^{(m)}_{\textrm{strength}}/z'$}, where {\small$z' = \mathbb{E}_{m'}[I^{(m')}_{\textrm{strength}}]$} was used for normalization. The classification of more texture categories made the DNN encode fewer middle-order interactions. \emph{I.e.} the stricter encoding of fine-grained textures usually led to fewer middle-order interactions.

$\bullet\quad$\textbf{Understanding 1-flexibility.}
\emph{If training samples contain noises, then the strength of low/middle-order interactions learned for shapes is relatively less flexible than the strength of low/middle-order interactions learned for textures. I.e. the encoding of textures is more flexible than the encoding of shapes.}
\begin{figure}[t]
	\centering
	\includegraphics[width=0.99\linewidth]{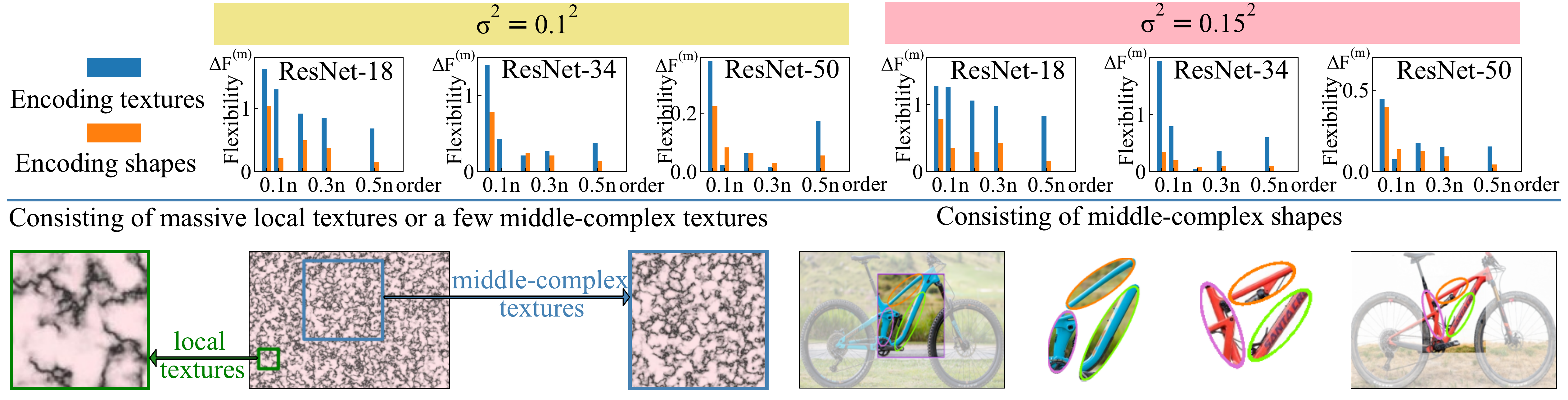}
	\vspace{-10pt}
	\caption{Flexibility of the $m$-order interaction strength $\Delta F^{(m)}$.} 
	\label{noise}
	\vspace{-13pt}
\end{figure}

This can be understood as follows.
Local simple repetitive textures are usually encoded as low-order interactions. The middle-order interaction for textures may be encoded in two different manners. \emph{I.e.} the middle-order interaction can either be modeled as the collaboration of simple/elementary textures corresponding to low-order interactions, or be memorized as a larger-scale middle-complex textural patch in a pixel-wise manner.
For example, the texture of the marble in Fig.~\ref{noise} is composed of repetitive local textures. In this way, this texture can be encoded as either the ensemble of massive local textures or the ensemble of a few middle-complex textures.
In comparison, for shapes, low-order interactions often represent local shapes. Middle-order interactions mainly represent collaborations between \emph{middle-level} features of different local shapes, instead of \emph{directly} memorizing the middle-complex shape in a \emph{pixel-wise} manner due to the shape deformation. 
For example, the DNN encodes a bicycle as the ensemble of middle-complex compositional parts (\emph{e.g.} the frame in Fig.~\ref{noise}), instead of massive local shapes.
Thus, the encoding of textures is more flexible.

\textbf{\emph{Verification.}} 
Here, we considered the flexibility of encoding textures and shapes when we learned DNNs under various noisy conditions. If DNNs learned under different noisy conditions had similar distributions of the multi-order interaction strength, we considered the learning of concepts was not flexible; otherwise, the learning of concepts was flexible. To this end, we aimed to verify the encoding of shapes was usually less flexible than the encoding of textures.

Thus, for encoding shapes, we trained DNNs using $30$ categories with the largest numbers of samples in the Caltech 101 dataset~\cite{fei2004learning} for object classification. Meanwhile, we used the aforementioned dataset with 66 texture categories, which was constructed based on the KTH-TIPS-2b dataset, to train DNNs for texture classification. Moreover, we added Gaussian noises with zero mean and variance {\small$\sigma^{2} \in \{0.1^{2}, 0.15^{2}\}$} to each training sample to train ResNet-18/34/50 models for comparisons.

In Fig.~\ref{noise}, we computed {\small$\Delta F^{(m)} \overset{\textrm{def}}{=} |F^{(m), \textrm{noise}}-F^{(m)}|$} to reflect the flexibility of multi-order interactions. 
{\small$F^{(m), \textrm{noise}}$} and {\small$F^{(m)}$} denoted the interaction strength of the DNN learned with noises and the DNN learned without noise, respectively. Note that $F^{(m), \textrm{noise}}$ and $F^{(m)}$ had been normalized for fair comparisons as aforementioned.
In this way, a large $\Delta F^{(m)}$ value indicated the $m$-order interaction strength was flexible under noisy conditions. 
Fig.~\ref{noise} shows the encoding of textures was more flexible than the encoding of shapes.

$\bullet\quad$\textbf{Understanding 1-generalization.}
 \emph{ 
Compared to middle-order interactions, low-order interactions extracted from the same image have similar utilities in classification, i.e. low-order interactions push the influence towards the same category.
Therefore, low-order interactions are usually learned to encode simple and common features of specific categories, just like bag-of-words models~\emph{\cite{harris1954distributional}}. For example, a local texture of the {\rm rape blossom} in Fig.~\ref{big} (\uppercase\expandafter{\romannumeral4}) is a common discriminative pattern for inference. In comparison, 
middle-order interactions extracted from the same image tend to represent more diverse, less common, and more complex concepts of various categories.}

\begin{figure*}[t]
\centering
\begin{minipage}[t]{0.47\linewidth}
	\centering
	\includegraphics[width=\linewidth]{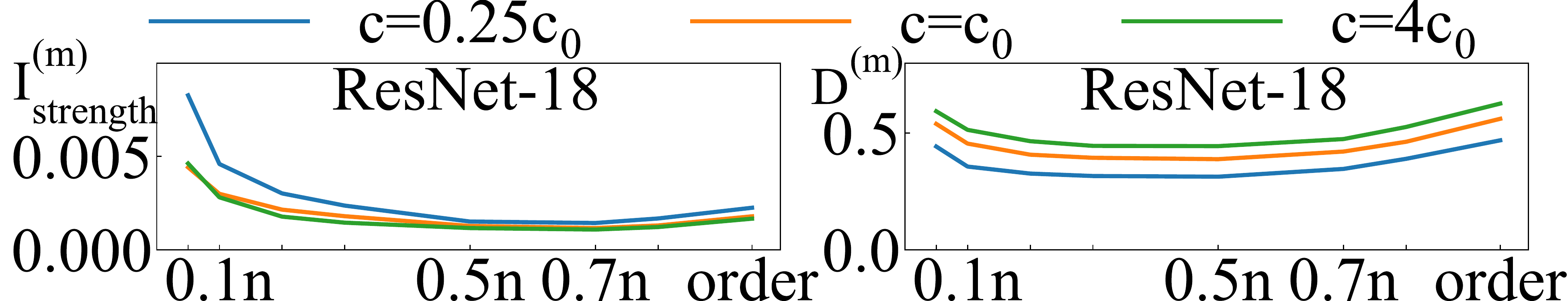}
	\vspace{-15pt}
	\caption{The $m$-order interaction strength {\small${I^{(m)}_{\textrm{strength}}}$} and disentanglement $D^{(m)}$ for ResNets with different layer widths, where $c_{0} = 64$ was the layer width the standard ResNet-18.}
	\label{channel}
	\vspace{-12pt}
\end{minipage}\;
\begin{minipage}[t]{0.5\linewidth}
\centering
	\includegraphics[width=\linewidth]{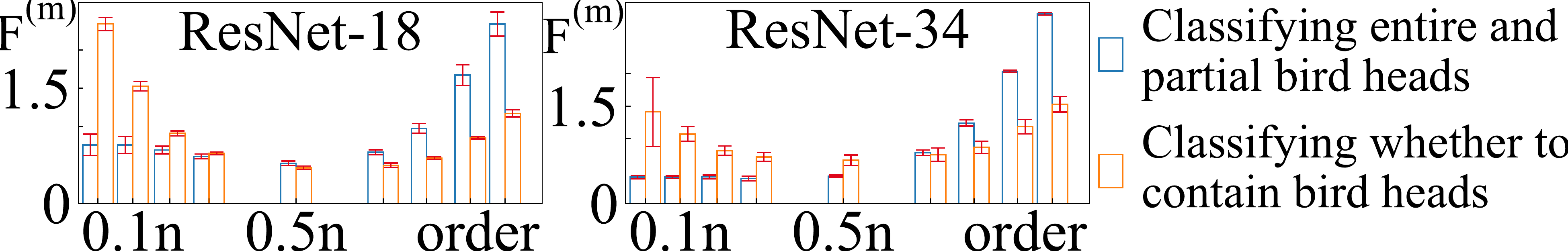}
	\vspace{-15pt}
	\caption{ $F^{(m)}$ for classifying entire and partial bird heads, and $F^{(m)}$ for classifying whether to contain bird heads, respectively. Red lines refer to standard deviations of results.}
	\label{parts_result}
	\vspace{-25pt}
\end{minipage}
\end{figure*}

\textbf{\emph{Verification.}} 
We proposed the disentanglement metric of the $m$-order interaction to check whether interactions extracted from the same image had similar utilities in classification.
\begin{equation}
		\begin{aligned}
		 D^{(m)} = \frac{ \mathbb{E}_{x\in \Omega}\mathbb{E}_{\substack{i,j\in N, i\ne j}}\big[|\sum_{S\subseteq N\setminus\{i,j\}, |S|=m} \Delta v(i,j,S|x)|\big]}{\mathbb{E}_{x\in \Omega}\mathbb{E}_{\substack{i,j\in N, i\ne j}}\big[ \sum_{S\subseteq N\setminus\{i,j\}, |S|=m} |\Delta v(i,j,S|x)|\big]},		
		\end{aligned}
\end{equation}
where the numerator equalled to {\small$I^{(m)}_{\textrm{strength}}$}. 
The disentanglement  {\small$D^{(m)}$} measured whether $m$-order interactions under different contexts had similar (either positive or negative) effects on the prediction. In this way,
a high disentanglement  {\small$D^{(m)}$} reflected that interactions between pixels $(i, j)$ under different contexts $\{S\}$ consistently had positive effects {\small$\Delta v(S,i,j|x)$} (or consistently had negative effects) on the inference of a specific category. Therefore, we could consider a large value of $D^{(m)}$ indicated that the $m$-order interaction mainly had similar classification utilities towards specific categories. Let us consider a toy example. When the pair of pixels $(i, j)$ consistently had a positive interaction towards a specific category,~\emph{i.e.} {\small$\forall S \in \{S\subseteq N\backslash \{i,j\}||S|=m\},\Delta v(S,i,j|x)>0$}, then we had  {\small$D^{(m)}=1$}. This indicated the $m$-order interaction between $i$ and $j$ stably promoted the output probability of this category. In contrast, a small value of {\small$D^{(m)}$} reflected interactions between $(i, j)$ referred to sophisticated concepts of various categories. \emph{I.e.} given different contexts, the interaction sometimes had positive effects on a specific category, and sometimes had negative effects.

Fig.~\ref{channel} shows low-order interactions encoded in the ResNet models with standard convolutional filters ($3 \times 3$) were more disentangled than middle-order interactions, which indicated low-order interactions were more likely to stably had similar classification utilities than middle-order interactions.
Experiments were introduced in Sec.~\ref{sec:4.4}, and supplementary materials also included more results computed using standard filters.

\subsection{Global concepts encoded as high-order interactions}
\begin{proposition} 
\label{pro2}
High-order interactions mainly represent the memory of specific large-scale concepts.
\end{proposition}\vspace{-5pt}
\begin{figure}[t]
	\centering
	\includegraphics[width=0.99\linewidth]{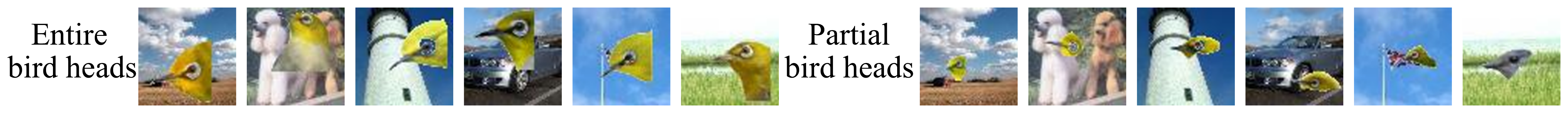}
	\vspace{-8pt}
	\caption{Positive training samples with entire bird heads and negative training samples with partial bird heads, which were used to verify \emph{understanding 2-semantics}.}
	\label{parts}
	\vspace{-18pt}
\end{figure}
$\bullet\quad$\textbf{Understanding 2-semantics.}
\emph{Unlike low/middle-order interactions mainly representing local/middle-complex concepts, high-order interactions mainly reflect specific large-scale visual concepts for inference. The large-scale (global) texture is a typical case. When a large-scale texture is encoded as a high-order interaction, it usually satisfies two requirements. 1) This large-scale textural concept frequently appears in images, e.g. the sky or the ocean. 2) When local textures are not discriminative enough, 
the DNN has to use high-order interactions over the entire image for inference. For example, in Fig.~\ref{big} (\uppercase\expandafter{\romannumeral2}), the background {\rm ocean} interacts with the foreground {\rm red-breasted merganser} for the classification of the {\rm red-breasted merganser}. The absence of either the background or the foreground will hurt the inference.}

\textbf{\emph{Verification 1.}} 
The basic idea was as follows. We constructed two datasets. One dataset forced the DNN to hard memorize the entire large-scale appearances of specific input instances for inference, while the other dataset did not. Thus, if the DNN had memorized these specific instances, it was supposed to encode more high-order interactions. 

Specifically, these two datasets were constructed using images from the Tiny ImageNet~\cite{le2015tiny}. To construct the first dataset, we pasted entire bird heads onto images as positive samples, and pasted partial bird heads to images as negative samples{\footnotemark[2]}, as Fig.~\ref{parts} shows. 
In this way, the DNN had to memorize the entire shape of the bird head for inference, which required complex collaborations with massive pixels. In comparison, to construct the second dataset, we pasted bird heads onto images as positive samples, and considered images without bird heads as negative samples{\footnotemark[2]}. In this case, partial information of bird heads, \emph{e.g.} beaks and eyes, was already discriminative enough to classify a bird head. It was not necessary for the DNN to memorize collaborations over the \emph{entire} bird head for classification. We repetitively trained multiple ResNet-18/34 models for classification 
with different parameter initialization states.
 Fig.~\ref{parts_result} shows the normalized interaction strength $F^{(m)}$ of all these DNNs. It was found that the DNN trained on the first dataset encoded more high-order interactions than the DNN trained on the second dataset.
Moreover, results in Fig.~\ref{parts_result} had small standard deviations, which proved the trustworthiness of our conclusion.

\textbf{\emph{Verification 2.}} 
In the supplementary material, we conducted the second experiment.
This experiment verified \emph{understanding 2-semantics} by the phenomenon that style classification (the style could be regarded as a large-scale texture) required more large-scale collaborations than object classification for inference.

$\bullet\quad$\textbf{Understanding 2-outlier.}
\emph{Like {\rm understanding 2-semantics}, the second typical case for high-order interactions is outlier samples.}

\textbf{\emph{Verification 1.}} 
 Because there was no standard way to define outlier samples in classification, we constructed different datasets with synthetic outliers. Specifically, we added negligible noises to some training samples, and assigned these noisy images with random labels to generate outliers. For verification, we proposed the purity metric \emph{w.r.t.} the $m$-order interaction, which examined whether most $m$-order interactions of different contexts contributed to the inference positively, with few $m$-order interactions having negative effects. Based on this purity metric, we found that high-order interactions were more useful for the classification of outliers than that of normal samples.
 
We constructed datasets with synthetic outlier samples based on the Tiny ImageNet dataset and the Caltech 101 dataset{\footnotemark[2]}, respectively. For the Tiny ImageNet, we added Gaussian noises with zero mean and variance $\sigma^{2}=0.05^{2}$ to 50/100/200/300 randomly chosen samples, and assigned these samples with random labels. Similarly, for the Caltech 101 dataset, we added noises to randomly selected 30/60/90 training samples ($\sigma^{2}=0.1^{2}$) and assigned them with random labels to construct three synthetic datasets. In this way, we trained ResNet-18 models on these datasets. Each DNN with high classification accuracy was supposed to memorize specific features of outlier samples, which required complex collaborations with massive pixels.  

For verification, we used two metrics to measure whether the classification of outliers more depended on high-order interactions than normal samples. The first metric was the average interaction {\small$I^{(m)}_{\textrm{avg}} = \mathbb{E}_{x \in \Omega}[\mathbb{E}_{i,j \in N}[I^{(m)}(i,j|x)]]$}, which reflected the average effect of $m$-order interactions for inference. We had proven that a large value of {\small$I^{(m)}_{\textrm{avg}}$} indicated that the $m$-order interaction made a significant contribution to the classification (proofs in the supplementary material). Another metric was the purity $P^{(m)}$, which measured the percentage of $m$-order interactions having positive effects among all $m$-order interactions,
\emph{i.e.}
{\small$P^{(m)} = \mathbb{E}_{x \in \Omega}\mathbb{E}_{i,j \in N}[\max (I^{(m)}(i,j|x),0)]/{I^{(m)}_{\textrm{strength}}}$.}
A large value of {\small$P^{(m)}$} indicated that more $m$-order interactions contributed to the classification positively, \emph{i.e.} being more useful. In this way, we computed the difference between outliers and normal samples, {\small$\Delta P^{(m)} \overset{\textrm{def}}{=} P^{(m), \textrm{outliers}} - P^{(m), \textrm{normal samples}}$, $\Delta I^{(m)}_{\textrm{avg}} \overset{\textrm{def}}{=}  I^{(m), \textrm{outliers}}_{\textrm{avg}} - I^{(m), \textrm{normal samples}}_{\textrm{avg}}$.} 

\begin{figure}[t]
	\centering
	\includegraphics[width=0.99\linewidth]{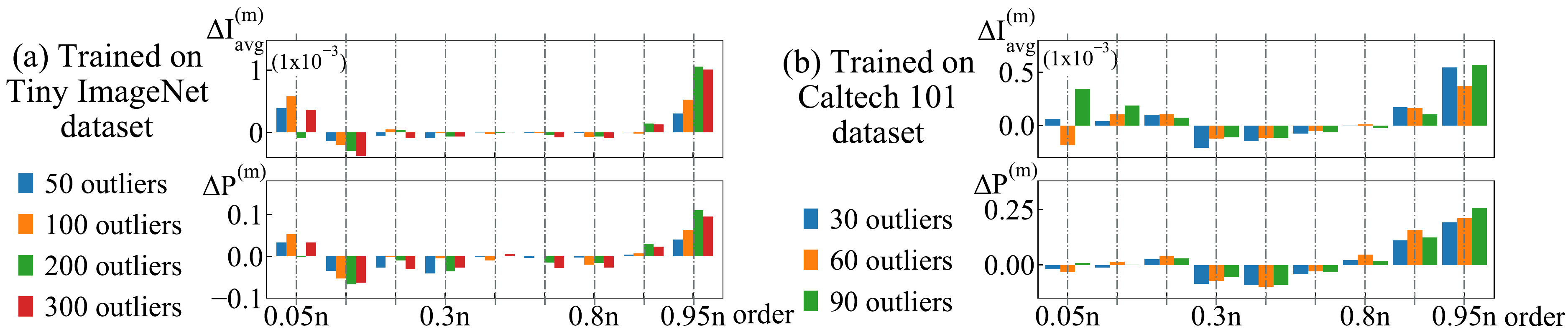}
	\vspace{-10pt}
	\caption{Differences between outlier samples and normal samples in the $m$-order average interaction $\Delta I^{(m)}_{\textrm{avg}}$ and the purity metric $\Delta P^{(m)}$, $m\in \{0.05n, 0.1n, 0.2n, 0.3n, 0.5n, 0.7n, 0.8n, 0.9n, 0.95n\}$.}
	\label{outlier}
	\vspace{-18pt}
\end{figure}

Fig.~\ref{outlier} shows {\small$\Delta P^{(m)}>0$} and {\small$\Delta I^{(m)}_{\textrm{avg}}>0$}, when $m>0.8n$.  It indicated that the classification of outliers mainly depended on high-order interactions.
In the supplementary material, given an LSTM model~\cite{hochreiter1997long} learned for sentiment classification using the SST-2 dataset~\cite{socher2013recursive}, we also found the phenomenon that outliers relied on high-order interactions for inference. 

\textbf{\emph{Verification 2.}} In the second experiment, \emph{understanding 2-outlier} was verified by the phenomenon that object classification of more categories (containing more hard samples, similar to outliers) encoded more high-order interactions for inference than object classification of fewer categories (see the supplementary material).

$\bullet\quad$\textbf{Understanding 2-complexity.}
\emph{
High-order interactions can be regarded as both complex and simple collaborations from different perspectives. 1) According to Eqn.~\eqref{eqn:multi-order_interaction}, high-order interactions are computed conditionally dependent on specific collaborations among most pixels in images. Such collaborations are complex and represent a specific input instance. 2) However, from another perspective, we can also consider high-order interactions as simple collaborations, which correspond to low-order interactions with the absence of a few pixels. I.e. high-order interactions measure how the absence of a few pixels destroys the global feature of an image. In this way, the DNN learns more ``{\rm simple}'' high-order interactions than middle-order interactions.
}

\textbf{\emph{Verification.}} Fig.~\ref{channel} and Fig.~\ref{kernel} verify that the strength of high-order interactions was larger than the strength of middle-order interactions. Experiments were introduced in Sec.~\ref{sec:4.4}.

\subsection{Combining taxonomy of concept complexities and taxonomy of textures and shapes}
Based on Propositions~\ref{pro1} and~\ref{pro2}, the new taxonomy of concept complexities provides a new perspective to re-classify textures and shapes encoded in DNNs into global (large-scale) textures, local textures, global (large-scale) shapes, and local shapes.
This helps people interpret the signal-processing behaviors of shapes and textures in a fine-grained manner.

To this end, we use the metric $\eta = \mathbb{E}_{i,j}[|I^{(m=0.9n)}(i,j|x)|] /z$ to describe high-order interactions, where $z = |v(N) - v(\emptyset)|$ is used for normalization. We separate images in the ImageNet validation dataset~\cite{russakovsky2015imagenet} into images with large $\eta$ values and images with small $\eta$ values. Among all images with small $\eta$ values, texture-biased images usually represent local textures, and shape-biased images often reflect local shapes. Similarly, among all images with large $\eta$ values, images containing large-scale texture-biased backgrounds are usually regarded as global textures, and images including large-scale rigid objects often represent global shapes. We use the method proposed by \citet{li2021shapetexture} to classify whether images are shape-biased or texture-biased.

\begin{figure*}[t]
	\centering
	\includegraphics[width=0.99\linewidth]{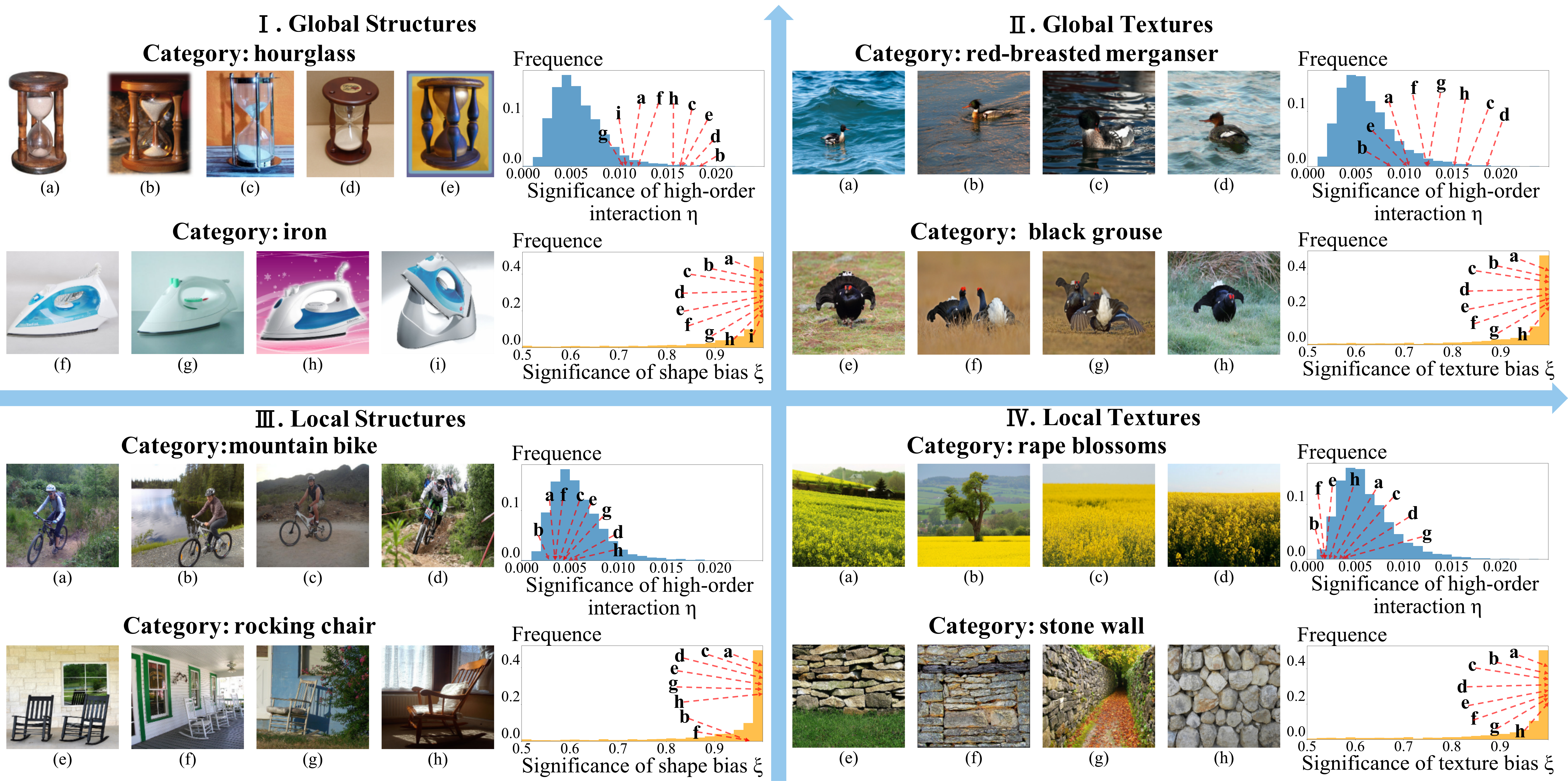}
	\vspace{-10pt}
	\caption{Typical samples of local textures, local shapes, global (large-scale) textures, and global (large-scale) shapes.}
	\label{big}
	\vspace{-15pt}
\end{figure*}

\emph{Type 1:} \emph{Hourglasses} and~\emph{irons} in Fig.~\ref{big} (\uppercase\expandafter{\romannumeral1}) usually exhibit strong high-order interactions, and are considered as global shapes. The common property of these images is that they all contain similar large-scale shapes without significant deformation. For example, samples of the hourglass category in the ImageNet dataset resemble each other. In this way, the global shape without significant deformation makes the DNN encode complex interactions between massive pixels as a specific feature for inference.

\emph{Type 2:} The \emph{red-breasted merganser} and the \emph{black grouse} with large texture-biased backgrounds in Fig.~\ref{big} (\uppercase\expandafter{\romannumeral2}) also exhibit strong high-order interactions, and are regarded as global textures. The common property of these images is that the classification depends on interactions between the foreground and the background, instead of exclusively relying on the foreground. For example, the absence of the \emph{ocean} background will hurt the inference of the \emph{red-breasted merganser}. Thus, the requirement for interactions between the foreground and background is the common cause for the DNN to encode global textures as high-order interactions.

\emph{Type 3:}~\emph{Mountain bikes} and~\emph{rocking chairs} shown in Fig.~\ref{big} (\uppercase\expandafter{\romannumeral3}) usually exhibit weak high-order interactions, and refer to local shapes. The common property for these images is that the classification mainly depends on local shapes of the foreground object. Although the background (\emph{e.g.} the \emph{grassland} or the \emph{lake} for \emph{mountain bikes}) can be regarded as textures to some extent, these textures are not discriminative for classification, thereby not being modeled by the DNN. 

\emph{Type 4:}~\emph{Rape blossoms} and \emph{stone walls} in Fig.~\ref{big} (\uppercase\expandafter{\romannumeral4}) with small $\eta$ values are encoded as local repetitive textures. The common property of such local textures is that they usually represent common simple repetitive textures. The local texture within larger contexts is already discriminative enough for inference. For example, simple collaborations among a local texture of \emph{rape blossoms} are sufficient for classification. Thus, it is not necessary for the DNN to encode complex collaborations with massive pixels for inference. In other words, the efficiency of local textures is a common reason for the DNN to encode them as low-order interactions.

\citet{li2021shapetexture} proposed a method to evaluate whether the DNN used textural information or shape information for inference. They used the classification confidence $\xi$ of a specific texture-biased model to examine whether textures were encoded for inference. Similarly, the classification confidence $\xi$ of a shape-biased model was used to check whether shapes were modeled for classification. In this way, we used $\xi$ values to distinguish texture-biased and shape-biased images, and used the normalized high-order interaction $\eta$ to distinguish global concepts and local concepts{\footnotemark[2]}.

Fig.~\ref{big} compares the frequency histogram of $\eta$ values and $\xi$ values over images in ImageNet, which were computed based on ResNet-50 models. We localized the selected images on each histogram, which illustrated the significance of the globality/locality and the shape/texture bias of images{\footnotemark[2]}. 

\subsection{Impacts of network architectures on interactions}
\label{sec:4.4}
\begin{figure}[t]
	\centering
	\includegraphics[width=0.77\linewidth]{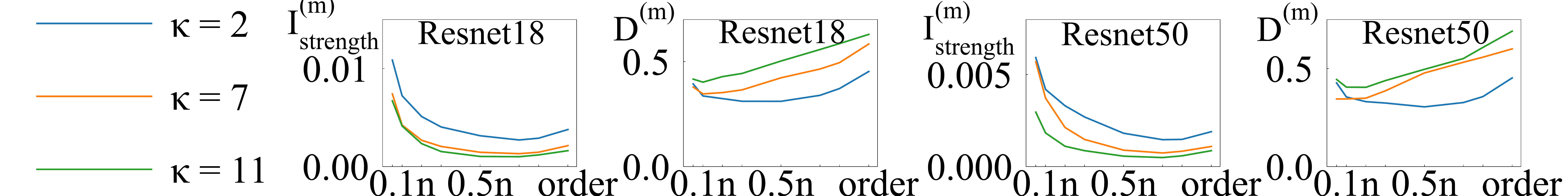}
	\vspace{-5pt}
	\caption{The $m$-order interaction strength ${I^{(m)}_{\textrm{strength}}}$ and the disentanglement $D^{(m)}$ for ResNets with different filter sizes $\kappa \times \kappa$.}
	\label{kernel}
	\vspace{-12pt}
\end{figure}

In this section, we further explored how the kernel size and the layer width of a ResNet influenced multi-order interactions. Given a ResNet, we constructed two baseline models with different kernel sizes. In order to simplify comparative studies, for each ResNet, we changed the kernel size of all $3 \times 3$ ($7 \times 7$) filters in convolutional layers to $\kappa \times \kappa$ filters, $\kappa \in \{2,7,11\}$. We used the Tiny ImageNet dataset to train baseline ResNet-18/50 models for object classification{\footnotemark[2]}.
Fig.~\ref{kernel} shows the ResNet with a larger kernel size often exhibited a smaller interaction strength {\small$I^{(m)}_{\textrm{strength}}$} and a higher disentanglement {\small$D^{(m)}$}. It indicated the DNN with a large kernel size was more likely to learn a few interactions of similar classification utilities (\emph{i.e.} learning more discriminative and more reliable interactions,
see \emph{understanding 1-generalization}), which encoded global collaborations with more pixels. This could be considered as the memory of a few complex but reliable interactions. In comparison, the DNN with a small kernel size tended to encode plenty of widely-shared, generic, but less discriminative interactions for classification.

Similarly, given a ResNet, we constructed baseline models with different layer widths (channel numbers). We used convolutional layers of widths $[c, 2c, 4c, 8c]$ for $c \in \{0.25c_{0}, c_{0}, 4c_{0}\}$, where $c_{0} = 64$ corresponded to the standard ResNet-18. We also used the Tiny ImageNet dataset to train baseline ResNet-18 models for object classification{\footnotemark[2]}.
Fig.~\ref{channel} shows a wider ResNet usually exhibited a smaller interaction strength and a higher disentanglement. It indicated a wide ResNet tended to encode a few discriminative interactions towards a specific category. Whereas, a narrow ResNet did not have enough filters, so it had to learn generic interactions shared by various categories. Nevertheless, we will conduct experiments on DNNs with more diverse architectures in the future.

Note that the phenomenon, \emph{i.e.} the middle order interaction had a smaller disentanglement than the low-order interaction, only existed when the DNN used standard (small) filters (shown in Fig.~\ref{channel}).
It was because small filters usually encouraged the feature sharing of low-order interactions. However, when the DNN used extremely large filters (\emph{e.g.} $11\times 11$ or $7 \times 7$), higher-order interactions were more disentangled (shown in Fig.~\ref{kernel}). It was because these large filters naturally encouraged the memory of specific instances.

\subsection{Clarifying practical values of multi-order interactions}
In this study, we have proven that the multi-order interaction can successfully reflect the complexity of visual concepts encoded in a DNN. In this section, we further discuss recent achievements of using multi-order interactions to improve the network performance. In this way, we can associate visual concepts of different complexities with deep-learning utilities.

$\bullet\quad$\textbf{Penalizing high-order interactions to improve the classification performance.}
Both \emph{understanding 2-outlier} and findings in \cite{zhang2021interpreting} had well shown that high-order interactions usually were responsible for the overfitting problem of a DNN. \citet{zhang2021interpreting} also proposed a method to improve the classification performance by penalizing the interaction strength.

$\bullet\quad$\textbf{{Explaining and alleviating adversarial effects.}}
\citet{ren2021game} discovered that adversarial perturbations usually affected high-order interactions.
Then, the study \cite{ren2021game} also explained the detection of adversarial samples \cite{yang2020ml} as the detection of salient high-order interactions.
Besides, many defense methods \cite{devries2017improved,jere2020singular,yang2020ml} were explained by \cite{ren2021game} as the removal of high-order interactions.

\section{Conclusion}
In this paper, we use multi-order interactions to rethink how a DNN encodes visual concepts of different complexities and explain the signal-processing behaviors of shapes and textures in a fine-grained manner. In this way, we classify textures into three types based on the multi-order interaction. The modeling of textures and that of shapes are distinguishable from the perspective of interactions. Besides, we use multi-order interactions to explain how a DNN encodes outliers. We also analyze impacts of network architectures on interactions modeled by the DNN, and clarify the crucial role of multi-order interactions in real-world applications. 
In conclusion, our study provides a new perspective to classify and analyze concepts of different complexities, beyond the categorical taxonomy of objects and the cognitive taxonomy of textures and shapes.

\bibliography{semantics}
\bibliographystyle{plainnat}
\end{document}